\documentclass{article} 
\usepackage[preprint]{iclr2027_conference_preprint_centered}
\usepackage{times}

\usepackage{amsmath,amsfonts,bm}

\def\eqref#1{equation~\ref{#1}}

\def\1{\bm{1}}

\DeclareMathAlphabet{\mathsfit}{\encodingdefault}{\sfdefault}{m}{sl}
\SetMathAlphabet{\mathsfit}{bold}{\encodingdefault}{\sfdefault}{bx}{n}

\usepackage[hidelinks]{hyperref}
\usepackage{url}

\usepackage{makecell}
\usepackage{float}
\usepackage{bm}
\usepackage{graphicx}
\usepackage{booktabs}
\usepackage{multirow}
\usepackage{enumitem}
\setlist[itemize]{leftmargin=*}
\setlist[enumerate]{leftmargin=*}

\usepackage{colortbl}
\usepackage{color}
\usepackage[table]{xcolor}
\usepackage[normalem]{ulem}
\useunder{\uline}{\ul}{}
\usepackage{subfigure}
\usepackage{wrapfig}
\usepackage{amsmath}
\usepackage{algorithm}
\usepackage{algpseudocode}

\title{Trading Strategy Optimization via Textual Gradient}

\author{Chaoqun Yang\qquad Qian Wang\qquad Fengbin Zhu\thanks{Corresponding author}\qquad Xinyu Lin \\
\textbf{Bingsheng He\qquad Roger Zimmermann\qquad Tat-Seng Chua} \\
National University of Singapore \\
\texttt{chaoqun@yang.email.cn\quad zhfengbin@gmail.com}
}

\begin{document}

\maketitle

\begin{abstract}
Quantitative trading strategy design aims to discover trading programs from historical data that remain effective in future markets, which can be viewed as a black-box program optimization problem. LLM-based textual gradients offer a promising approach by providing explicit optimization directions for iterative strategy refinement. However, directly applying textual gradients faces two challenges: (1) optimization is myopic, underutilizing experience from previous evaluations; and (2) aggregate backtest feedback overlooks temporal robustness, potentially favoring strategies that perform well only in specific market periods. To address these challenges, we propose TradeGrad, an experience-guided textual-gradient framework for robust trading strategy optimization. TradeGrad leverages accumulated optimization experience to estimate textual gradients and employs multi-scale revisions for both strategy exploration and refinement. It further introduces the Cross-Period Robust Objective (CPRO), which emphasizes performance in unfavorable historical periods to promote temporal robustness. Experiments on cross-sectional and time-series strategy design in Chinese A-share and U.S. equity markets show that TradeGrad achieves the best in-sample and out-of-sample performance across all four settings. Notably, its Chinese cross-sectional strategy achieves 27.99\% annualized return, 12.19\% maximum drawdown, and a Sharpe ratio of 1.63, approximately 68\% higher than the CSI 300 benchmark. Further analyses validate the proposed components and show consistent improvements in both in-sample and out-of-sample performance throughout optimization. The code is available at \url{https://github.com/transcend-0/TradeGrad}.
\end{abstract}
\section{Introduction}

\begin{figure}[!t]
    \setlength{\abovecaptionskip}{0cm}
    \setlength{\belowcaptionskip}{0cm}
    \centering
    \includegraphics[width=1\linewidth]{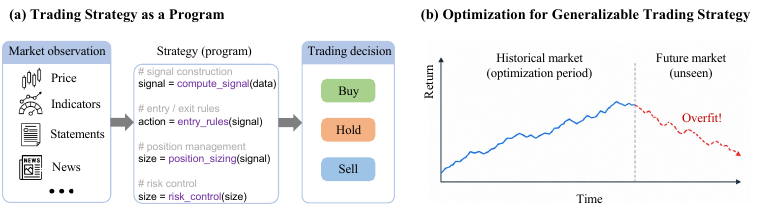}
    \vspace{-5mm}
    \caption{(a) Trading Strategy as a Program. (b) Optimization for Generalizable Trading Strategy.}
    \label{fig:intro}
    \vspace{-5mm}
\end{figure}

Quantitative trading strategies are systematic decision rules that map market observations into trading decisions and play an important role in modern investment~\citep{guo2024quant}. In practice, a strategy is implemented as an executable program that integrates signal construction, entry and exit rules, position management, and risk control (Figure~\ref{fig:intro}(a)). 
Trading strategy design can therefore be viewed as searching over a vast space of programs~\citep{allen1999using}. 
However, unlike conventional program search under a fixed evaluation setting, trading strategy design uses historical market data to discover programs for deployment in future markets, where conditions may differ substantially. 
The objective is thus to identify strategies whose effectiveness extends beyond historical backtests to unseen future market conditions~\citep{bailey2014deflated} (Figure~\ref{fig:intro}(b)).

Traditionally, trading strategies were designed by quantitative researchers through a combination of financial expertise, domain intuition, and extensive empirical experimentation. The substantial expertise and effort required have motivated research into automated strategy design~\citep{allen1999using,lo2000foundations}.
Recent advances in Large Language Models (LLMs) offer new opportunities to automate this process by combining code generation with evolutionary search~\citep{novikov2025alphaevolve,lange2026shinkaevolve,yun2025quantevolve}.
Leveraging their prior knowledge of programming and finance, LLMs can generate various strategy code and thereby facilitate the exploration of the strategy space.
Nevertheless, existing methods rely largely on stochastic mutations, providing limited guidance on how a strategy should be improved and thus requiring extensive search to discover effective refinements.

Textual gradient optimization~\citep{yuksekgonul2025optimizing} offers a more direct paradigm for iterative program refinement. Analogous to numerical gradients, it is defined as a natural-language optimization signal that diagnoses weaknesses in the current program and suggests actionable directions for revision. For example, if a momentum strategy performs well in trending markets but suffers large drawdowns during volatile reversals, a textual gradient may suggest normalizing momentum by recent volatility or requiring stronger trend consistency before entering a position. The LLM can then translate this semantic optimization direction into concrete modifications of the strategy program.

However, directly applying textual gradients to trading strategy optimization is unsatisfactory due to two key limitations.
\textbf{(1) Textual-gradient optimization is myopic.} A gradient inferred from the current strategy captures only a narrow view of the strategy space. Earlier successes and failures are easily forgotten, causing redundant exploration and unstable optimization. Moreover, a fixed scale of revision cannot meet the varying needs of the search process: exploring alternative trading hypotheses requires structural changes, whereas refining promising strategies calls for localized edits.
\textbf{(2) Aggregate backtest feedback overlooks temporal robustness.} Financial markets are non-stationary, and strategy performance can vary substantially across historical periods~\citep{bailey2014deflated,bailey2017probability,arian2024backtest}. An aggregate backtest score can be dominated by strong performance in favorable periods, even when the strategy performs poorly in others. Textual gradients derived from such feedback therefore direct optimization toward period-specific improvements that fail to generalize across market regimes. Effective optimization requires feedback that explicitly accounts for performance across different historical periods.

To address the above limitations, we propose \textbf{TradeGrad}, an experience-guided textual-gradient framework for robust trading strategy optimization, built around two central ideas.
\textbf{(1) Experience-guided textual-gradient estimation} accumulates and reuses knowledge across strategy revisions. TradeGrad maintains a global optimization memory that distills successful and failure patterns from evaluated strategies. To estimate textual gradients, it samples archived strategies with their code and backtest results, providing comparative evidence for effective modifications. TradeGrad further employs multi-scale revisions: structural revisions explore alternative trading hypotheses, while localized code edits refine promising strategies.
\textbf{(2) Cross-Period Robust Objective (CPRO)} promotes temporal robustness by evaluating strategies across calendar-year windows. Within each window, CPRO computes a composite score that integrates annualized return, maximum drawdown, and a one-sided penalty for insufficient trading activity. Inspired by Conditional Value-at-Risk (CVaR)~\citep{rockafellar2002conditional}, CPRO aggregates the lower tail of these window-level scores, emphasizing unfavorable periods rather than allowing a few favorable periods to dominate.

We evaluate TradeGrad on cross-sectional and time-series strategy design in both Chinese A-share and U.S. equity markets. 
Extensive experimental results demonstrate that TradeGrad not only consistently achieves the highest in-sample (IS) and out-of-sample (OOS) scores among the evaluated strategy-design methods across all four settings but also maintains favorable risk–return trade-offs.
Notably, in the Chinese market, its cross-sectional strategy achieves an annualized return of $27.99\%$ with a maximum drawdown of only $12.19\%$ and a Sharpe ratio of $1.63$, outperforming the CSI 300 index by about $68\%$ in Sharpe ratio. Ablation studies confirm that all proposed components are essential to TradeGrad's optimization performance. 
Optimization trajectory analysis shows that TradeGrad consistently improves both IS and OOS performance throughout the optimization, whereas compared methods tend to plateau or deteriorate.

The key contributions of this work are summarized as follows:
\begin{itemize}
\item We formulate trading strategy design as a black-box program optimization problem, and identify two limitations of applying textual-gradient optimization in this setting: myopic optimization that makes insufficient use of accumulated search experience, and aggregate backtest feedback that overlooks temporal robustness.
\item We propose TradeGrad, an experience-guided textual-gradient framework for robust trading strategy optimization. By combining global optimization memory, sampling-based gradient estimation, and multi-scale strategy revisions, TradeGrad enables more efficient strategy optimization.
\item TradeGrad introduces the Cross-Period Robust Objective that evaluates strategies across multiple calendar-year windows and emphasizes lower-tail performance, thereby promoting improvements that remain effective across market periods and generalize to unseen markets.
\item Extensive experiments on cross-sectional and time-series strategy design in Chinese A-share and U.S. equity markets demonstrate that TradeGrad achieves the best IS and OOS performance among all evaluated methods and maintains favorable risk–return trade-offs. Ablation studies, hyperparameter analyses, and optimization trajectory analysis provide further insights into TradeGrad's effectiveness.
\end{itemize}

\section{Problem Formulation}

\subsection{Black-Box Optimization over Strategy Programs}

We formulate automated trading strategy design as a black-box optimization problem over executable programs.
Let $\Pi$ denote the space of executable strategy programs, and let $\mathcal{D}=\{x_t\}_{t=1}^{T}$ denote a sequence of market observations, where $x_t$ contains the information available at time $t$. A trading strategy is represented by an executable program $\pi \in \Pi$ that maps the observed market history to a trading decision $a_t=\pi(x_{\leq t})$. A strategy may jointly specify signal construction, entry and exit rules, position management, portfolio allocation, and risk control.

Given a strategy $\pi$ and a market dataset $\mathcal{D}$, a predefined backtesting procedure produces a backtest result $\mathcal{B}(\pi;\mathcal{D})$, which contains performance statistics such as annualized return, maximum drawdown, Sharpe ratio, and trading frequency. The quality of a strategy is evaluated through an objective function
$J(\mathcal{B}) = J(\pi;\mathcal{D})$.
In general, strategy evaluation may involve multiple objectives reflecting different aspects of trading performance. For notational simplicity, we use a scalar $J$ throughout the paper to represent the objective used by the optimizer, where multiple performance criteria can be aggregated into $J$ when necessary.

The optimizer iteratively proposes new strategy programs based on previously evaluated candidates. Specifically, a generic black-box optimizer can be represented as a proposal operator $\mathcal{U}$, which generates the strategy at iteration $k$ as
$\pi_{k} = \mathcal{U}\left(\{(\pi_i,\mathcal{B}_i,J_i)\}_{i=0}^{k-1}\right)$,
where $\mathcal{B}_i$ denotes the backtest result of strategy $\pi_i$, and $J_i$ denotes its corresponding objective value.


\subsection{Optimization for Generalizable Trading Strategy}

A distinctive challenge in trading strategy optimization is that performance on the historical data available during optimization is not itself the ultimate target. The goal is to obtain a strategy that performs well on future and unseen markets. Let $p_{\mathrm{future}}$ denote the unknown distribution of future market data. Ideally, the strategy should maximize
\begin{equation}
    \pi^*
    =
    \arg\max_{\pi\in\Pi}
    \mathbb{E}_{\mathcal{D}\sim p_{\mathrm{future}}}
    \left[
        J(\pi;\mathcal{D})
    \right].
\end{equation}

In practice, $p_{\mathrm{future}}$ is unavailable during strategy development. We therefore divide the available historical data into an IS period $\mathcal{D}_{\mathrm{IS}}$ and an OOS period $\mathcal{D}_{\mathrm{OOS}}$. During optimization, the optimizer can access only backtest evaluations on $\mathcal{D}_{\mathrm{IS}}$, whereas $\mathcal{D}_{\mathrm{OOS}}$ is held out from the optimization process and used only to evaluate whether the resulting strategies generalize to unseen market periods.

\section{Method}

\subsection{Overview}

\begin{figure}[t]
    \setlength{\abovecaptionskip}{0cm}
    \setlength{\belowcaptionskip}{0cm}
    \centering
    \includegraphics[width=1\linewidth]{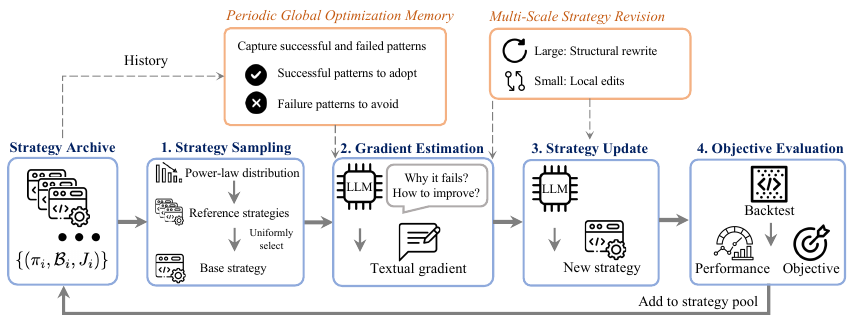}
    \vspace{-4mm}
    \caption{
        Overview of TradeGrad. 
        TradeGrad iteratively samples archived strategies, estimates textual gradients from backtest feedback and optimization experience, updates strategies at varying revision scales, and evaluates and archives the resulting strategies.
    }
    \label{fig:method}
    \vspace{-3mm}
\end{figure}

We propose TradeGrad, a textual-gradient optimization framework for automated trading strategy design, as illustrated in Figure~\ref{fig:method} and Algorithm~\ref{alg:method}. 
TradeGrad maintains an archive of evaluated strategies and reuses them as optimization experience. At each iteration, it samples several strategies and selects one as the base strategy. An LLM compares the sampled strategies and their backtest results, together with experience summarized from previous iterations, to identify improvements to the base strategy. The resulting textual gradient guides its revision, ranging from structural rewrites to local edits. Finally, the new strategy is evaluated across multiple historical periods and added back to the archive, forming an iterative optimization loop.

\subsection{Experience-Guided Textual Gradient Optimization}

\subsubsection{Periodic Global Optimization Memory}

As optimization proceeds, evaluated strategies and their backtest outcomes accumulate in the archive $\mathcal{A}_k$. To avoid repeatedly providing the growing archive to the LLM, TradeGrad maintains a persistent global memory $\mathcal{M}_k$ that periodically distills recent optimization experience into reusable successful and failure patterns. Formally, every $H$ iterations, the preceding $H$ experiences are summarized and consolidated with the previous memory:
\begin{equation}
    \mathcal{M}_k
    =
    S_{\mathrm{global}}
    \left(
        \left\{
        S_{\mathrm{local}}(\pi_i,\mathcal{B}_i,J_i)
        \right\}_{i=k-H}^{k-1},
        \mathcal{M}_{k-H}
    \right),
    \qquad
    k\bmod H=0.
    \label{eq:global_memory}
\end{equation}
Here, $S_{\mathrm{local}}$ extracts reusable successful and failure patterns from an individual optimization experience, while $S_{\mathrm{global}}$ consolidates these summaries with the existing memory. 
Between refreshes, the latest memory is retained. 
The resulting $\mathcal{M}_k$ provides persistent experience that complements the strategies sampled from the current archive.

\subsubsection{Sampling-Based Textual Gradient Estimation}

At iteration $k$, strategies in $\mathcal{A}_{k-1}$ are ranked by their objective values $J$. We sample $M$ reference strategies without replacement according to the power-law distribution \begin{equation} 
P(\pi_{(r)}) = \frac{r^{-\alpha_s}} {\sum_{j=1}^{|\mathcal{A}_{k-1}|}j^{-\alpha_s}}, 
\label{eq:reference_sampling} 
\end{equation} 
where $\pi_{(r)}$ denotes the strategy with rank $r$ and $\alpha_s>0$ controls the preference for highly ranked strategies. 
This produces a performance-biased reference set 
$\mathcal{S}_k=\{(\pi_k^{(j)},\mathcal{B}_k^{(j)},J_k^{(j)})\}_{j=1}^{M}$. 
We then uniformly select one strategy from $\mathcal{S}_k$ as the base strategy $\pi_k^{\mathrm{base}}$. 
Thus, the reference context is biased toward stronger strategies, while the update target can originate from different quality levels. 

Given the reference set $\mathcal{S}_k$, base strategy $\pi_k^{\mathrm{base}}$, global memory $\mathcal{M}_k$, and revision scale $\eta_k$, the LLM estimates a textual gradient 
\begin{equation} 
\hat{g}_k = \mathrm{LLM} \left( \mathcal{S}_k, \pi_k^{\mathrm{base}}, \mathcal{M}_k, \eta_k \right).
\label{eq:textual_gradient} 
\end{equation} 
The textual gradient identifies weaknesses of the base strategy and specifies actionable modifications by comparing strategy programs and their backtest outcomes.

\subsubsection{Multi-Scale Strategy Revision}

Program revisions can differ substantially in semantic magnitude, from restructuring the core strategy logic to modifying a local trading condition. To balance these behaviors, we sample a discrete revision scale
\begin{equation}
    \eta_k
    =
    \begin{cases}
        \eta_{\mathrm{large}}, & \text{with probability } p_{\mathrm{large}},\\
        \eta_{\mathrm{small}}, & \text{with probability } 1-p_{\mathrm{large}}.
    \end{cases}
\end{equation}

The revision scale $\eta_k$ specifies the intended scope of modification during textual gradient generation and subsequently determines how the program is revised. When $\eta_k=\eta_{\mathrm{large}}$, the LLM is instructed to make substantial changes to the core trading logic, and the resulting strategy is implemented as a full rewrite of the base program. When $\eta_k=\eta_{\mathrm{small}}$, the LLM focuses on localized improvements while preserving the overall strategy structure, and the revision is implemented through diff-based replacements of selected code segments. Therefore, large-scale revisions promote structural exploration, while small-scale revisions enable targeted refinement.

\subsection{Cross-Period Robust Objective}
\label{sec:objective}

\subsubsection{Cross-Period Evaluation}

Aggregate backtest performance may favor strategies whose gains are concentrated in a few favorable periods. Therefore, we evaluate each strategy across multiple historical windows. Specifically, we partition the in-sample data into $N$ non-overlapping historical windows,
$\mathcal{D}_{\mathrm{IS}} = \left\{ \mathcal{D}^{(1)},\ldots,\mathcal{D}^{(N)} \right\}$.
For each window, annualized return $AR_n(\pi)$ and maximum drawdown $MDD_n(\pi)$ are transformed into normalized scores:
\begin{align}
    s_{\mathrm{AR}}^{(n)}(\pi)
    &=
    \sigma\!\left(
        \lambda_{\mathrm{AR}}
        (AR_n(\pi)-r_{\mathrm{ref}})
    \right),
    \\
    s_{\mathrm{MDD}}^{(n)}(\pi)
    &=
    \sigma\!\left(
        -\lambda_{\mathrm{MDD}}
        (MDD_n(\pi)-d_{\mathrm{ref}})
    \right),
    \label{eq:period_scores}
\end{align}
where $\sigma(x)=1/(1+e^{-x})$ is the sigmoid function, $\lambda_{\mathrm{AR}}$ and $\lambda_{\mathrm{MDD}}$ control the sensitivities, and $r_{\mathrm{ref}}$ and $d_{\mathrm{ref}}$ are reference levels for return and drawdown, respectively. This transformation maps the two metrics to a common bounded scale while favoring higher returns and lower drawdowns.

We combine the normalized scores with a one-sided trading-frequency gate:
\begin{equation}
    s_n(\pi)
    =
    100 \times
    \left(s_{\mathrm{AR}}^{(n)}(\pi)\right)^{\alpha}
    \left(s_{\mathrm{MDD}}^{(n)}(\pi)\right)^{1-\alpha}
    \left[
        \min\left(
            1,
            \frac{\tau_n(\pi)}{\tau_0}
        \right)
    \right]^{\gamma},
    \label{eq:window_score}
\end{equation}
where $\alpha$ controls the trade-off between return and drawdown, $\tau_n(\pi)$ is the trading frequency, $\tau_0$ is the desired minimum frequency, and $\gamma$ controls the penalty strength. The gate penalizes insufficient trading activity without rewarding excessive trading beyond $\tau_0$.

\subsubsection{Lower-Tail Robust Objective}

Inspired by Conditional Value-at-Risk (CVaR)~\citep{rockafellar2002conditional}, we focus on the lower tail of window-level performance to emphasize unfavorable historical periods. Specifically, we sort the window scores as
$s_{[1]}(\pi) \leq \cdots \leq s_{[N]}(\pi)$
and optimize the mean of the lowest $30\%$:
\begin{equation}
    J(\pi;\mathcal{D}_{\mathrm{IS}})
    =
    \frac{1}{q}
    \sum_{n=1}^{q}s_{[n]}(\pi),
    \qquad
    q=\left\lceil0.3N\right\rceil.
    \label{eq:robust_objective}
\end{equation}
We refer to this objective as the Cross-Period Robust Objective (CPRO). By emphasizing lower-tail performance, CPRO prevents strong results in favorable periods from fully compensating for poor performance elsewhere, thereby promoting robustness across market periods.

After generating $\pi_k$, TradeGrad evaluates it using Equation~(\ref{eq:robust_objective}) and adds the strategy and its evaluation to the archive, which provides experience for subsequent textual-gradient updates.
\section{Experiments}

In this section, we conduct extensive experiments to answer the following research questions: \textbf{RQ1}: Does TradeGrad discover better trading strategies than existing methods? 
\textbf{RQ2}: How do the proposed mechanisms contribute to performance? 
\textbf{RQ3}: How do different hyperparameters affect the performance? 
\textbf{RQ4}: How does the optimization evolve throughout the search process?

\subsection{Experimental Setup}
\label{sec:experimental_setup}

\textbf{Markets and Evaluation.}
We evaluate TradeGrad on two representative markets: the Chinese A-share market and the U.S. stock market. For both markets, we use daily market data from 2018 to 2025, chronologically split into an IS optimization period and an OOS evaluation period with a ratio of $3{:}1$. 
We consider two representative strategy-design settings. 
The first is the cross-sectional setting, where strategies trade individual stocks and construct portfolios by comparing assets at the same point in time. 
We use the point-in-time constituents of the CSI 300 for the Chinese market and the S\&P 500 for the U.S. market.
The second is the time-series setting, where strategies trade the corresponding market index based on its historical temporal dynamics.
We evaluate strategies using IS Score (the value of the CPRO), OOS Score (the value of the aggregate objective), Annualized Return (AR), Maximum Drawdown (MDD), and Sharpe Ratio (SR). More implementation details are provided in Appendix~\ref{app:experimental_setup}.

\textbf{Baselines.}
We compare TradeGrad against four groups of baselines:
\textbf{(1) Market index.} We report the corresponding market index as a passive benchmark for each market.
\textbf{(2) Classical strategies.} We consider Momentum~\citep{jegadeesh1993returns}, Reversal~\citep{de1985does}, Volatility~\citep{ang2006cross}, and Volume~\citep{gervais2001high} for cross-sectional trading, and Grid Trading~\citep{rundo2019grid}, Moving Average (MA)~\citep{brock1992simple}, Channel Breakout~\citep{brock1992simple}, and Bollinger Bands~\citep{bollinger2002bollinger} for time-series trading.
\textbf{(3) Evolutionary program search.} We consider two representative baselines. The first is the Island Algorithm, which combines LLM-driven code evolution with an island-based evolutionary framework and underlies general-purpose program evolution methods such as AlphaEvolve~\citep{novikov2025alphaevolve} and ShinkaEvolve~\citep{lange2026shinkaevolve}. The second is QuantEvolve~\citep{yun2025quantevolve}, a domain-specific extension that incorporates financial guidance to tailor evolutionary search to quantitative trading.
\textbf{(4) Textual-gradient optimization.} We use TextGrad~\citep{yuksekgonul2025optimizing}, which iteratively updates a single strategy according to textual gradients derived from its evaluation feedback.

\subsection{Overall Performance (RQ1)}

\begin{table*}[t]
\centering
\caption{Overall performance comparison. Results are averaged over three runs; standard deviations are reported in Appendix~\ref{app:std}. Best and second-best results are bolded and underlined, respectively.}
\label{tab:main}
\setlength{\tabcolsep}{0.8mm}
\small
\begin{tabular}{llcccccccccc}
\toprule
& & \multicolumn{5}{c}{CSI 300} & \multicolumn{5}{c}{S\&P 500} \\
\cmidrule(lr){3-7}
\cmidrule(lr){8-12}
Setting & Method
& \shortstack{IS\\Score} $\uparrow$
& \shortstack{OOS\\Score} $\uparrow$
& AR $\uparrow$
& MDD $\downarrow$
& SR $\uparrow$
& \shortstack{IS\\Score} $\uparrow$
& \shortstack{OOS\\Score} $\uparrow$
& AR $\uparrow$
& MDD $\downarrow$
& SR $\uparrow$ \\
\midrule

\multicolumn{2}{l}{\textit{Market Index Reference}}
& 1.63 & 45.01 & 17.69 & 15.66 & 0.97
& 5.44 & 39.38 & 20.76 & 18.90 & 1.26 \\
\midrule

\multicolumn{12}{l}{\textit{Cross-Sectional Strategies}} \\

\multirow{4}{*}{Classical}
& Momentum
& 0.52 & 9.19 & 5.76 & 28.49 & 0.35
& 9.02 & 2.37 & -15.35 & 31.13 & -1.58 \\

& Reversal
& 0.28 & 14.51 & 17.80 & 28.59 & 0.74
& 5.12 & \underline{26.76} & \underline{0.56}
& \underline{11.59} & \underline{0.11} \\

& Volatility
& \underline{7.56} & \underline{60.69} & \underline{18.59}
& \textbf{8.74} & \underline{1.41}
& 4.41 & 6.93 & -9.38 & 23.79 & -0.79 \\

& Volume
& 1.51 & 16.73 & 6.92 & 23.03 & 0.41
& 13.62 & 2.43 & -16.08 & 30.54 & -2.57 \\

\cmidrule(lr){1-12}

\multirow{2}{*}{Evolution}
& Island Algorithm
& 0.23 & 2.52 & -12.74 & 31.87 & -0.60
& \underline{16.84} & 0.60 & -22.53 & 40.16 & -4.22 \\

& QuantEvolve
& 3.03 & 4.42 & -5.59 & 30.14 & -0.15
& 0.34 & 0.16 & -26.75 & 49.90 & -2.30 \\

\cmidrule(lr){1-12}

\multirow{2}{*}{Gradient}
& TextGrad
& 0.09 & 4.40 & -8.08 & 28.94 & -0.40
& 1.02 & 3.51 & -14.92 & 27.63 & -2.62 \\

& \cellcolor{gray!20}\textbf{TradeGrad}
& \cellcolor{gray!20}\textbf{10.69}
& \cellcolor{gray!20}\textbf{65.65}
& \cellcolor{gray!20}\textbf{27.99}
& \cellcolor{gray!20}\underline{12.19}
& \cellcolor{gray!20}\textbf{1.63}
& \cellcolor{gray!20}\textbf{21.36}
& \cellcolor{gray!20}\textbf{40.38}
& \cellcolor{gray!20}\textbf{6.05}
& \cellcolor{gray!20}\textbf{3.93}
& \cellcolor{gray!20}\textbf{1.07} \\

\midrule

\multicolumn{12}{l}{\textit{Time-Series Strategies}} \\

\multirow{4}{*}{Classical}
& Grid
& 4.32 & \underline{35.52} & \underline{6.34} & 11.30 & \underline{0.73}
& 7.03 & 7.59 & -11.39 & 21.66 & -1.21 \\

& MA
& 12.48 & 17.55 & -0.22 & 18.52 & 0.04
& 0.67 & 10.08 & \textbf{24.09} & 33.54 & 0.88 \\

& Breakout
& 25.69 & 28.11 & 1.13 & 10.96 & 0.17
& 4.60 & 30.42 & 2.64 & 10.77 & 0.36 \\

& Bollinger
& 22.81 & 13.63 & -7.89 & 16.50 & -1.36
& 5.13 & 3.80 & -14.37 & 27.13 & -1.28 \\

\cmidrule(lr){1-12}

\multirow{2}{*}{Evolution}
& Island Algorithm
& \underline{26.04} & 30.08 & 3.50 & 12.26 & 0.38
& 14.22 & 36.02 & 11.57 & 15.77 & 1.06 \\

& QuantEvolve
& 9.73 & 24.10 & 2.94 & 16.14 & 0.30
& \underline{20.18} & \underline{45.40} & 9.64
& \underline{7.10} & \underline{1.09} \\

\cmidrule(lr){1-12}

\multirow{2}{*}{Gradient}
& TextGrad
& 0.89 & 29.48 & 0.05 & \textbf{6.49} & 0.03
& 0.65 & 32.93 & 2.23 & \textbf{6.18} & 0.34 \\

& \cellcolor{gray!20}\textbf{TradeGrad}
& \cellcolor{gray!20}\textbf{29.00}
& \cellcolor{gray!20}\textbf{41.54}
& \cellcolor{gray!20}\textbf{8.21}
& \cellcolor{gray!20}\underline{8.59}
& \cellcolor{gray!20}\textbf{0.87}
& \cellcolor{gray!20}\textbf{23.24}
& \cellcolor{gray!20}\textbf{50.08}
& \cellcolor{gray!20}\underline{12.39}
& \cellcolor{gray!20}7.79
& \cellcolor{gray!20}\textbf{1.32} \\

\bottomrule
\end{tabular}
\vspace{-5mm}
\end{table*}

We first answer RQ1 by evaluating the IS and OOS performance of the designed strategies. From the results shown in Table~\ref{tab:main}, we have the following findings.


\textbf{Classical strategies remain competitive, while automated baselines struggle with cross-sectional trading.}
Volatility ranks second in OOS Score on CSI 300 cross-sectional trading, and Grid outperforms all automated baselines on CSI 300 time-series trading. In contrast, Island Algorithm, QuantEvolve, and TextGrad incur OOS losses in both cross-sectional markets. These automated baselines fare better in time-series trading, all achieving positive OOS returns in both markets, with QuantEvolve ranking second in OOS Score on S\&P 500.

\textbf{TradeGrad combines effective IS optimization with strong OOS generalization.}
TradeGrad achieves the highest IS and OOS Scores among all compared strategy-design methods in all four settings. This alignment distinguishes it from baselines whose historical success fails to carry over: For example, Island Algorithm ranks second in IS Score on S\&P 500 cross-sectional trading but nearly last in OOS Score. TradeGrad's consistently leading performance on both splits supports its ability to discover strong historical strategies whose advantage persists on unseen data.

\textbf{TradeGrad achieves favorable OOS risk--return trade-offs.}
TradeGrad obtains the highest OOS Sharpe ratio among strategy-design methods in all four settings. On CSI 300 cross-sectional trading, it achieves the highest AR of $27.99\%$ with MDD of $12.19\%$. On S\&P 500 cross-sectional trading, it delivers a positive AR of $6.05\%$ with the lowest MDD of $3.93\%$. Its advantage also extends beyond maximizing returns: on S\&P 500 time-series trading, TradeGrad earns a lower AR than MA but reduces MDD from $33.54\%$ to $7.79\%$ and increases SR from $0.88$ to $1.32$.

\subsection{Ablation Study (RQ2)}

\subsubsection{Ablation of Experience-Guided Textual-Gradient Estimation}

\begin{table}[t]
\centering
\caption{Ablation of experience-guided textual-gradient estimation. Results are mean $\pm$ standard deviation over three runs. Best and second-best means are bolded and underlined, respectively.}
\label{tab:optimization_ablation}
\small
\begin{tabular}{lccccc}
\toprule
Method & IS Score $\uparrow$ & OOS Score $\uparrow$ & AR $\uparrow$ & MDD $\downarrow$ & SR $\uparrow$ \\
\midrule
\rowcolor{gray!20}
\textbf{Full Method}
& $\mathbf{10.69}_{\pm 1.06}$
& $\mathbf{65.65}_{\pm 12.15}$
& $\mathbf{27.99}_{\pm 4.19}$
& $12.19_{\pm 4.65}$
& $\mathbf{1.63}_{\pm 0.17}$ \\

w/o Memory
& $4.56_{\pm 0.42}$
& $35.62_{\pm 7.07}$
& $11.78_{\pm 1.73}$
& $16.09_{\pm 5.97}$
& $0.80_{\pm 0.08}$ \\

w/o Sampling
& $\underline{8.32}_{\pm 0.85}$
& $46.72_{\pm 10.68}$
& $12.98_{\pm 2.03}$
& $\underline{11.35}_{\pm 4.38}$
& $0.94_{\pm 0.10}$ \\

w/o MSR (Diff)
& $5.79_{\pm 0.54}$
& $46.88_{\pm 12.10}$
& $14.87_{\pm 1.14}$
& $13.04_{\pm 5.14}$
& $0.96_{\pm 0.10}$ \\

w/o MSR (Rewrite)
& $7.42_{\pm 0.77}$
& $\underline{55.74}_{\pm 10.01}$
& $\underline{15.98}_{\pm 2.39}$
& $\mathbf{8.86}_{\pm 3.48}$
& $\underline{0.98}_{\pm 0.11}$ \\

\bottomrule
\end{tabular}
\vspace{-5mm}
\end{table}

To answer RQ2, we first ablate the three textual-gradient optimization mechanisms in Table~\ref{tab:optimization_ablation}: periodic global memory, sampling-based gradient estimation, and multi-scale revision (MSR). Under the same CPRO and evaluation budget, we remove memory or sampling, and replace MSR with Diff-only or Rewrite-only updates.

\textbf{All three optimization mechanisms contribute to the final performance.}
Removing any proposed mechanism decreases both IS and OOS Scores. The full method achieves an OOS Score of $65.65$, while removing sampling or memory and replacing MSR with a single revision mode all lead to substantial degradation. This consistent decline indicates that the three mechanisms provide complementary benefits to textual-gradient optimization.

\textbf{Global memory makes the largest contribution to optimization.}
Removing global memory causes the largest performance drop among all ablations, reducing the OOS Score from $65.65$ to $35.62$. This suggests that retaining optimization experience across iterations is critical for guiding textual-gradient updates toward strategies that generalize beyond the optimization period.

\textbf{MSR balances local refinement and broader exploration.}
Both Diff-only and Rewrite-only revisions underperform the full method, with Rewrite-only performing better than Diff-only. This suggests that broader revisions help escape weak search regions, while combining them with fine-grained modifications enables MSR to balance exploration and refinement more effectively.

\subsubsection{Ablation of the Cross-Period Robust Objective}

\begin{table}[t]
\centering
\caption{Ablation of the CPRO. IS OV reports each variant's own objective value and is not directly comparable across rows. Results are mean $\pm$ standard deviation over three runs. Best and second-best means are bolded and underlined, respectively.}
\label{tab:objective_ablation}
\setlength{\tabcolsep}{1.2mm}
\small
\begin{tabular}{lcccccc}
\toprule
Objective & IS OV $\uparrow$ & IS Score $\uparrow$ & OOS Score $\uparrow$ & AR $\uparrow$ & MDD $\downarrow$ & SR $\uparrow$ \\
\midrule

\rowcolor{gray!20}
\textbf{CPRO}
& $10.69_{\pm 1.06}$
& $\mathbf{10.69}_{\pm 1.06}$
& $\mathbf{65.65}_{\pm 12.15}$
& $\mathbf{27.99}_{\pm 4.19}$
& $\mathbf{12.19}_{\pm 4.65}$
& $\mathbf{1.63}_{\pm 0.17}$ \\

CPRO w/o Freq Gate
& $10.99_{\pm 1.07}$
& $\underline{8.10}_{\pm 0.79}$
& $\underline{40.24}_{\pm 8.68}$
& $10.43_{\pm 1.51}$
& $\underline{12.66}_{\pm 4.81}$
& $0.73_{\pm 0.07}$ \\

Mean-Period Objective
& $37.72_{\pm 4.18}$
& $4.40_{\pm 0.40}$
& $23.37_{\pm 5.06}$
& $6.08_{\pm 0.93}$
& $18.70_{\pm 7.13}$
& $0.44_{\pm 0.05}$ \\

Aggregate Objective
& $5.11_{\pm 0.56}$
& $0.74_{\pm 0.07}$
& $35.63_{\pm 7.64}$
& $\underline{13.12}_{\pm 1.96}$
& $16.89_{\pm 6.38}$
& $\underline{0.77}_{\pm 0.09}$ \\

\bottomrule
\end{tabular}
\vspace{-3mm}
\end{table}

We further examine the effect of the optimization objective on OOS generalization in Table~\ref{tab:objective_ablation}, comparing CPRO with variants that remove the frequency gate, average performance across periods, or evaluate performance over the aggregated optimization period.

\textbf{The trading-frequency gate improves both IS performance and OOS generalization.}
Removing the frequency gate reduces the standardized IS Score from $10.69$ to $8.10$ and the OOS Score from $65.65$ to $40.24$. It also lowers OOS profitability and increases drawdown, with SR declining from $1.63$ to $0.73$. These results support the gate's role in guiding search toward strategies that perform better under the common evaluation criteria and retain their advantage out of sample.

\textbf{Emphasizing weak periods is more effective than optimizing mean or aggregate performance.}
CPRO achieves the highest standardized IS Score, compared with $4.40$ for the Mean-Period Objective and $0.74$ for the Aggregate Objective. Its advantage extends to OOS evaluation, where it achieves a Score of $65.65$, versus $23.37$ and $35.63$, respectively, together with higher AR and SR and lower MDD. These results support emphasizing the lower tail of historical performance: under the same search procedure and budget, this objective produces strategies with stronger performance under both the common IS criterion and held-out OOS evaluation.

\subsection{Hyperparameter Analysis (RQ3)}

\begin{figure}[t]
    \setlength{\abovecaptionskip}{0cm}
    \setlength{\belowcaptionskip}{0cm}
    \centering
    \includegraphics[width=1\linewidth]{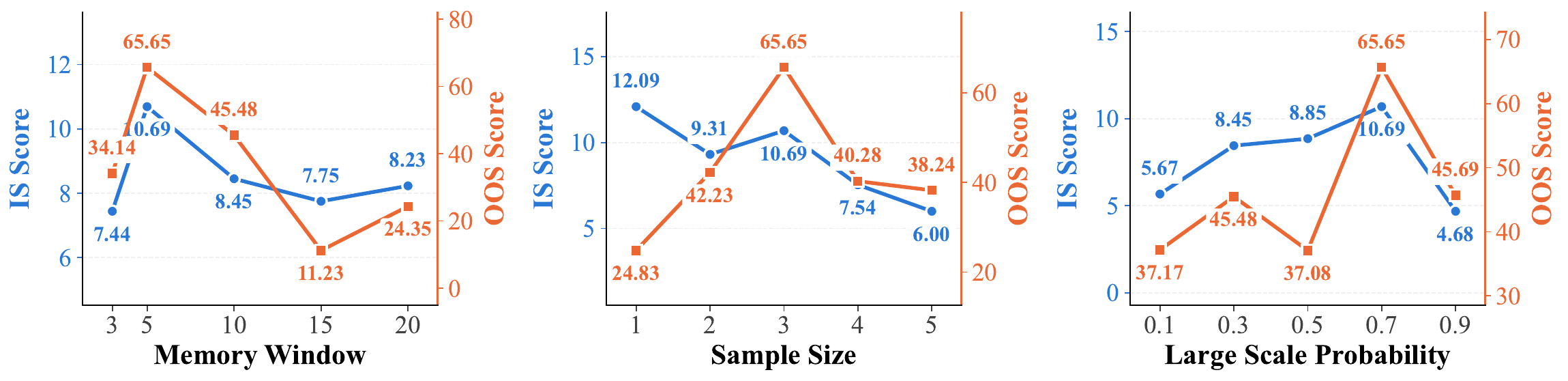}
    \vspace{-5mm}
    \caption{Effect of different hyperparameters: Memory window for summarizing optimization experience, sample size for textual-gradient estimation, and probability of large-scale revision.}
    \label{fig:hyperparameter}
    \vspace{-3mm}
\end{figure}


 


Figure~\ref{fig:hyperparameter} presents a sensitivity analysis of three key hyperparameters, varying one at a time while keeping the others fixed. The analysis is intended to characterize the sensitivity of TradeGrad and to examine whether its behavior is consistent across the optimization and held-out periods.

\textbf{Global memory exhibits a non-monotonic sensitivity to its temporal horizon.}
Performance improves as the memory window increases from $3$ to $5$, but generally declines with longer windows. One possible explanation is that longer windows may incorporate stale or heterogeneous experience that weakens subsequent updates.
 
\textbf{The sample size shows different IS and OOS trends.}
A single sample achieves the highest IS Score, whereas a sample size of $3$ shows stronger OOS performance. Further increasing the sample size degrades both metrics, suggesting that additional references may provide useful comparative evidence but can also introduce irrelevant or conflicting information.

\textbf{MSR benefits from a mixture of revision scales.}
OOS performance generally improves as the probability of large revisions increases, but drops substantially at $p_{\mathrm{large}}=0.9$. Together with the MSR ablation, this pattern is consistent with the motivation for combining structural exploration with localized refinement.

\subsection{Optimization Trajectories (RQ4)}

\begin{figure}[t]
    \setlength{\abovecaptionskip}{0cm}
    \setlength{\belowcaptionskip}{0cm}
    \centering
    \includegraphics[width=1\linewidth]{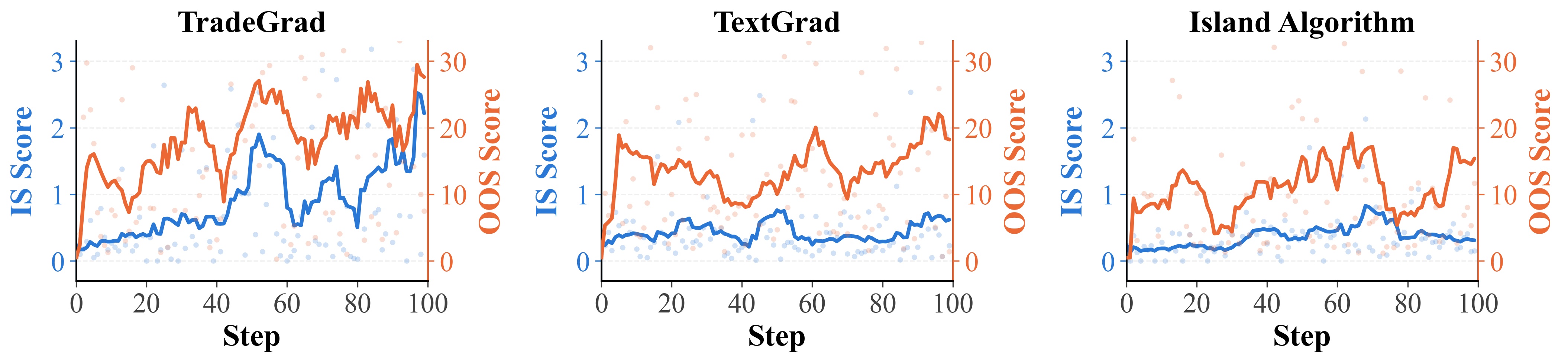}
    \vspace{-5mm}
    \caption{Optimization trajectories of different methods. Points denote the strategies generated at each optimization step, and curves show their 10-step moving averages (MA(10)).}
    \label{fig:trajectory}
\end{figure}

To further understand the search dynamics, we track the IS and OOS Scores of the strategies generated throughout optimization. Figure~\ref{fig:trajectory} shows the strategy obtained at each step together with its 10-step moving average (MA(10)), which highlights the overall optimization trend.

\textbf{TradeGrad achieves more sustained optimization and generalization than the baselines.}
TradeGrad shows a clear upward trend in both IS and OOS Scores throughout optimization, indicating that the search continues to discover stronger strategies and transfers these improvements out of sample. In contrast, TextGrad quickly plateaus after its initial gains, while Island Algorithm exhibits smaller and less persistent improvements with frequent regressions. Although all methods occasionally discover high-OOS strategies, the steadily rising trajectories of TradeGrad show that it more effectively accumulates optimization progress and translates it into sustained OOS gains. 
A case study illustrating the optimization process is provided in Appendix~\ref{app:case_study}.

\section{Related Work}

\textbf{Automated Trading Strategy Design.}
Automated trading strategy design aims to discover executable trading rules from market data with limited human specification. Early work primarily used genetic algorithms to evolve trading rules across equity, foreign-exchange, and futures markets, jointly searching over strategy structures and parameters~\citep{allen1999using,neely1999technical,wang2000trading,potvin2004generating,hu2015application}. Because flexible program search can overfit historical data, subsequent studies emphasize robustness through risk-aware objectives, transaction costs, out-of-sample evaluation, and evaluation across multiple market segments~\citep{neely2003risk,esfahanipour2011genetic,berutich2016robust}. More recently, LLM-GA~\citep{zhang2025llm}, QuantEvolve~\citep{yun2025quantevolve}, and ProFiT~\citep{siper2026profit} use LLMs to generate and refine executable strategies, but remain within the evolutionary-search paradigm. In contrast, our work explores textual-gradient optimization for trading strategy design, with an explicit focus on out-of-sample robustness.
More related work is discussed in Appendix~\ref{app:related_work}.

\section{Conclusion}

In this work, we formulate trading strategy design as a black-box program optimization problem and propose TradeGrad, an experience-guided textual-gradient framework for robust trading strategy optimization. TradeGrad leverages accumulated optimization experience to overcome the myopic nature of textual gradients and introduces the CPRO to promote temporal robustness across different market periods. Experiments on cross-sectional and time-series strategy design in both Chinese A-share and U.S. equity markets show that TradeGrad achieves strong in-sample and out-of-sample performance while maintaining favorable risk--return trade-offs, with further analyses validating the effectiveness of its key components. Future work can extend TradeGrad to broader asset classes and higher-frequency settings, incorporate richer market information into textual-gradient estimation, and investigate robustness under more diverse forms of distribution shift. More broadly, TradeGrad provides a promising direction for LLM-based program optimization in settings where solutions optimized on historical data need to generalize to unseen environments.
\subsection*{AI use statement}

In this work, we used generative AI tools for implementing methods. 
We have not used generative AI tools for generating synthetic datasets, formulating mathematical claims, proposing or refining hypotheses, designing or providing feedback on research methodology or experiments, assisting with translation, cleaning and reformatting datasets, supporting qualitative and thematic data analysis, interpreting results. Providing critical ingredients for proving mathematical claims and assisting in the writing of proofs are not applicable to this work.
Additionally, we used generative AI tools for creating or editing software code, drafting parts of a research paper, summarizing or analysing existing literature, sourcing/searching for information, editing a research paper to improve readability, and identifying relevant literature.
We have reviewed all AI-assisted work. In particular, all literature identified or listed with the assistance of generative AI tools was independently checked by the authors for accuracy and relevance. LLM-generated code was reviewed, tested, and modified by the authors to verify its correctness and ensure that it implemented the intended functionality.
We take responsibility for the final content of this work,
including text, claims or artifacts produced with the aid of generative AI.


\subsection*{Reproducibility statement}

We have taken several steps to ensure the reproducibility of our work. The source code is available on GitHub. All datasets used in this work are publicly accessible, with their sources and experimental splits described in Section~\ref{sec:experimental_setup}. We provide detailed implementation and evaluation settings in Appendix~\ref{app:experimental_setup}, including the evaluation metrics, TradeGrad hyperparameters and CPRO configuration, accessible market observations, optimization protocol, backtesting assumptions, and evaluation scope. Together, the released code, publicly available datasets, and documented experimental settings provide the necessary resources for reproducing the reported experiments and results.


\bibliography{main}
\bibliographystyle{iclr2027_conference}

\appendix
\section{Appendix}

\subsection{Algorithm}

The overall optimization procedure is summarized in Algorithm~\ref{alg:method}.

\begin{algorithm}[h]
\caption{\textbf{TradeGrad}}
\label{alg:method}
\begin{algorithmic}[1]
\Require Initial strategy $\pi_0$, maximum optimization steps $K$
\Ensure Optimized strategy $\pi^*$

\State $\mathcal{B}_0 \gets \mathcal{B}(\pi_0;\mathcal{D}_{\mathrm{IS}})$
\State $J_0 \gets J(\mathcal{B}_0)$
\State $\mathcal{A} \gets \{(\pi_0,\mathcal{B}_0,J_0)\}$
\State $\mathcal{M} \gets \emptyset$

\For{$k=1,\ldots,K$}
    \If{$k\bmod H = 0$}
        \State $\mathcal{W}
            \gets
            \{(\pi_i,\mathcal{B}_i,J_i)\}_{i=k-H}^{k-1}$
        \State $\mathcal{M}
            \gets
            \textsc{UpdateMemory}(\mathcal{W},\mathcal{M})$
    \EndIf

    \State $\mathcal{S}_k
        \sim
        \textsc{RankBasedSample}(\mathcal{A})$
    \State $\pi_k^{\mathrm{base}}
        \sim
        \textsc{UniformSample}(\mathcal{S}_k)$
    \State $\eta_k
        \gets
        \textsc{SelectLR}()$

    \State $\hat{g}_k
        \gets
        \textsc{TextGradient}
        (\mathcal{S}_k,\pi_k^{\mathrm{base}},\mathcal{M},\eta_k)$

    \State $\pi_{k}
        \gets
        \textsc{Update}
        (\pi_k^{\mathrm{base}},\hat{g}_k,\eta_k)$

    \State $\mathcal{B}_k
        \gets
        \mathcal{B}
        (\pi_k;\mathcal{D}_{\mathrm{IS}})$
    \State $J_k
        \gets
        J(\mathcal{B}_k)$
    \State $\mathcal{A}
        \gets
        \mathcal{A}
        \cup
        \{(\pi_k,\mathcal{B}_k,J_k)\}$

\EndFor

\State $\pi^*
    \gets
    \arg\max_{(\pi,\mathcal{B},J)\in\mathcal{A}} J$
\State \Return $\pi^*$

\end{algorithmic}
\end{algorithm}

\subsection{Implementation Details}
\label{app:experimental_setup}

\subsubsection{Evaluation Metrics}

For in-sample evaluation, to ensure a consistent basis for comparison, we report the value of the CPRO as the IS Score for all methods. For automatically searched strategies, the trading-frequency gate is included when computing the reported IS Score, whereas for classical strategies and the market-index reference, the frequency penalty is omitted.

For the optimization signal, following common practice in trading strategy optimization, the classical strategies and automated baselines are optimized or selected according to IS SR, except QuantEvolve, which follows the objective specified in its original method~\citep{yun2025quantevolve}.
TradeGrad instead uses CPRO for both optimization and strategy selection.
Thus, the reported IS Score provides a common evaluation criterion while preserving the original optimization protocol of each method.

For out-of-sample evaluation, after strategy selection is completed using IS data, the selected strategy is evaluated once on the strictly held-out OOS period.
We compute the OOS Score over the entire OOS period as
\begin{equation}
    \mathrm{OOS\ Score}(\pi)
    =
    100
    \left(
        s_{\mathrm{AR}}^{\mathrm{OOS}}(\pi)
    \right)^{\alpha}
    \left(
        s_{\mathrm{MDD}}^{\mathrm{OOS}}(\pi)
    \right)^{1-\alpha},
    \label{eq:oos_score}
\end{equation}
where $s_{\mathrm{AR}}^{\mathrm{OOS}}(\pi)$ and
$s_{\mathrm{MDD}}^{\mathrm{OOS}}(\pi)$ are computed from the annualized return and maximum drawdown over the entire OOS period using the transformations in Equation~(\ref{eq:period_scores}).
Unlike IS Score, the OOS Score does not perform cross-period lower-tail aggregation or apply the trading-frequency gate.

We additionally report three standard OOS performance metrics: 
\begin{itemize}
    \item Annualized Return (AR): annualized growth of portfolio value.
    \item Maximum Drawdown (MDD): The largest peak-to-trough decline in cumulative returns.
    \item Sharpe Ratio (SR): Risk-adjusted return relative to return volatility.
\end{itemize}



\subsubsection{Experimental Protocol}

\textbf{Market Observations.}
At each trading step, a strategy can access up to the preceding 252 trading days of historical market observations. 
For the Chinese A-share market, the available daily fields are open, high, low, close, volume, amount, and turnover. 
For the U.S. market, the available daily fields are open, high, low, close, and volume.
No observations from the held-out OOS period are exposed to the optimizer during strategy optimization or selection.

\textbf{Optimization Protocol.}
All LLM-based methods use Gemini-3-Flash\footnote{\url{https://storage.googleapis.com/deepmind-media/Model-Cards/Gemini-3-Flash-Model-Card.pdf}}. 
To ensure comparability, all LLM-based methods use the same strategy representation, initial strategy information, accessible market observations, backtesting environment, and transaction-cost assumptions. 
Each method is given a maximum budget of 100 candidate strategies. We use the number of generated candidate strategies as the primary search budget because backtesting each candidate constitutes the common black-box evaluation operation across methods.

\textbf{Backtesting Assumptions.}
All backtests start with an initial capital of $1{,}000{,}000$. 
We apply a proportional slippage rate of $0.05\%$ and a commission rate of $0.05\%$ per transaction. 
For the Chinese A-share market, we additionally apply a stamp tax rate of $0.1\%$ on applicable transactions. 
All transaction costs are directly incorporated into strategy returns during backtesting.
Chinese-market strategies are restricted to long-only positions, whereas U.S.-market strategies can take both long and short positions.

\textbf{Evaluation Scope.}
The main experiments cover both markets and both strategy settings. All ablation studies, hyperparameter analyses, and optimization trajectory analyses are conducted in the CSI 300 cross-sectional setting.
All reported results are averaged over three independent runs. For readability, Table~\ref{tab:main} reports the mean values, while the corresponding standard deviations are
provided in Appendix~\ref{app:std}.

\subsection{Standard Deviations of Main Results}
\label{app:std}

To assess the robustness of the experimental results, we repeat each stochastic method three times with different random seeds. Table~\ref{tab:std} reports the standard deviations corresponding to the mean results in Table~\ref{tab:main}. Overall, the results exhibit consistent performance
across independent runs.

\begin{table*}[t]
\centering
\caption{Standard deviations of the main results over three independent runs.}
\label{tab:std}
\setlength{\tabcolsep}{1.5mm}
\small
\begin{tabular}{lcccccccccc}
\toprule
& \multicolumn{5}{c}{CSI 300} & \multicolumn{5}{c}{S\&P 500} \\
\cmidrule(lr){2-6}
\cmidrule(lr){7-11}
Method
& \shortstack{IS\\Score}
& \shortstack{OOS\\Score}
& AR
& MDD
& SR
& \shortstack{IS\\Score}
& \shortstack{OOS\\Score}
& AR
& MDD
& SR \\
\midrule

\multicolumn{11}{l}{\textit{Cross-Sectional Strategies}} \\

Island Algorithm
& 0.02 & 0.52 & 2.01 & 11.26 & 0.06
& 1.12 & 0.13 & 3.37 & 15.03 & 0.44 \\

QuantEvolve
& 0.31 & 0.94 & 0.84 & 11.04 & 0.02
& 0.03 & 0.03 & 4.02 & 18.54 & 0.24 \\

Vanilla TextGrad
& 0.01 & 0.97 & 1.21 & 11.50 & 0.04
& 0.10 & 0.74 & 2.23 & 10.09 & 0.27 \\

\rowcolor{gray!20}
\textbf{Ours}
& 1.06 & 12.15 & 4.19 & 4.65 & 0.17
& 2.67 & 8.39 & 0.94 & 1.50 & 0.11 \\

\midrule

\multicolumn{11}{l}{\textit{Time-Series Strategies}} \\

Island Algorithm
& 2.56 & 4.49 & 0.53 & 4.86 & 0.04
& 1.40 & 7.79 & 1.74 & 6.02 & 0.12 \\

QuantEvolve
& 0.98 & 5.13 & 0.44 & 6.26 & 0.03
& 2.01 & 9.76 & 1.34 & 2.97 & 0.11 \\

Vanilla TextGrad
& 0.09 & 6.45 & 0.01 & 2.48 & 0.01
& 0.06 & 7.10 & 0.33 & 2.36 & 0.04 \\

\rowcolor{gray!20}
\textbf{Ours}
& 2.90 & 8.86 & 1.32 & 3.24 & 0.08
& 2.32 & 10.70 & 1.86 & 2.71 & 0.14 \\

\bottomrule
\end{tabular}
\end{table*}

\subsection{Case Study}
\label{app:case_study}

\begin{figure}[t]
\setlength{\abovecaptionskip}{0cm}
\setlength{\belowcaptionskip}{0cm}
\centering
\includegraphics[width=1\linewidth]{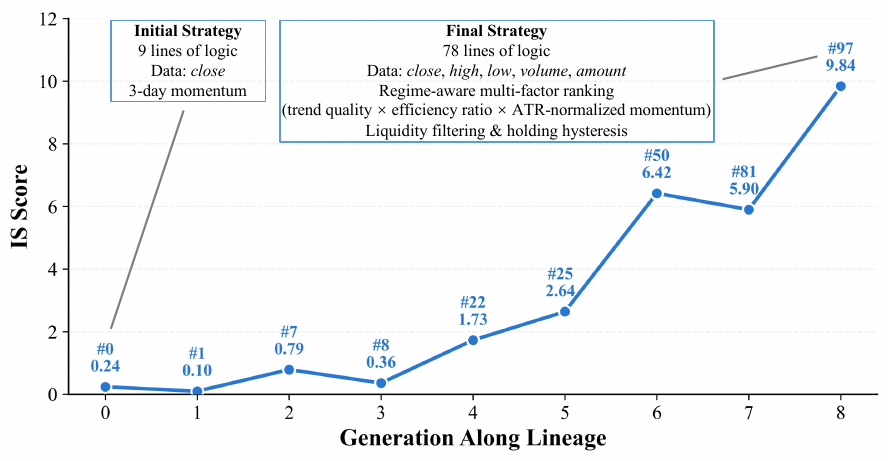}
\vspace{-5mm}
\caption{Evolution of a final strategy along its ancestral lineage. Labels indicate program IDs and IS Scores.}
\label{fig:case_study}
\end{figure}


To understand how TradeGrad transforms trading programs during optimization, we trace the ancestral lineage of a representative final strategy. Figure~\ref{fig:case_study} shows how the strategy logic and IS Score evolve from the initial program to the final solution.

\paragraph{TradeGrad evolves simple heuristics into substantially richer trading programs.}
The initial strategy contains only nine lines of logic, using three-day momentum computed solely from closing prices. Through successive revisions, TradeGrad expands both the information set and the decision structure. The final strategy contains 78 lines of logic and incorporates price, volume, and trading-amount information. More importantly, it evolves from a single momentum signal into a regime-aware multi-factor strategy that combines trend quality, efficiency ratio, and ATR-normalized momentum with liquidity filtering and holding hysteresis. This evolution shows that TradeGrad searches beyond parameter tuning or local signal adjustment, progressively restructuring the executable trading logic.

\paragraph{Strategy evolution is non-monotonic but accumulates into substantial improvement.}
The ancestral lineage does not follow monotonic hill climbing. Several revisions temporarily reduce the IS Score, including $0.24 \rightarrow 0.10$, $0.79 \rightarrow 0.36$, and $6.42 \rightarrow 5.90$, yet these programs remain on the lineage leading to stronger descendants. Across eight generations, the strategy ultimately evolves from an IS Score of $0.24$ to $9.84$. This case illustrates the value of maintaining flexible program exploration: revisions that are not immediately beneficial can introduce new logic that becomes useful after subsequent refinement, allowing TradeGrad to reach solutions that would be difficult to obtain through strictly greedy updates.

\subsection{More Related Work}
\label{app:related_work}

\textbf{LLM-based Program Optimization.}
LLMs have recently emerged as powerful operators for black-box program optimization, enabling search directly over executable code. FunSearch~\citep{romera2024mathematical} combines LLM-based program generation, automated evaluation, and evolutionary selection, a paradigm subsequently extended to reward functions, heuristics, and metaheuristics by Eureka~\citep{ma2024eureka}, EoH~\citep{liu2024evolution}, ReEvo~\citep{ye2024reevo}, and LLaMEA~\citep{van2024llamea}, and further scaled or improved in AlphaEvolve~\citep{novikov2025alphaevolve} and ShinkaEvolve~\citep{lange2026shinkaevolve}. These methods primarily optimize programs through evolutionary search, whereas textual gradients provide explicit semantic directions for program revision. Moreover, unlike typical program-optimization settings that optimize performance on a given evaluation problem, trading strategies must generalize from historical optimization data to unseen markets; our work studies textual-gradient program optimization under this additional generalization requirement.

\textbf{Textual Gradient Optimization.}
Textual-gradient optimization uses natural-language feedback as an optimization signal for discrete and non-differentiable objects. Prior work has largely focused on prompt optimization: ProTeGi~\citep{pryzant2023automatic} uses LLM-generated critiques to guide revisions, OPRO~\citep{yang2024large} conditions optimization on previously evaluated solutions, GPO~\citep{tang2025unleashing} studies gradient-inspired update directions and magnitudes, DLPO~\citep{peng2025dlpo} introduces textual learning rates and decay, and GEPA~\citep{agrawal2026gepa} combines natural-language reflection with evolutionary search. Beyond prompts, TextGrad~\citep{yuksekgonul2025optimizing} formalizes textual gradients that can propagate through computation graphs to optimize code and other structured objects, while TSGD-M~\citep{ding2026scaling} and Center-Aware Textual Gradients~\citep{jang2025prompt} improve their stability and reliability. Applying this paradigm to trading introduces two additional challenges: limited exploration from single-trajectory updates and overfitting to historical backtests. We address these challenges through population-based textual-gradient optimization and a cross-period robust objective.

\end{document}